\documentclass{article}

\usepackage[final]{neurips_2024_ml4ps}
\PassOptionsToPackage{numeric}{natbib}
\usepackage[utf8]{inputenc} 
\usepackage[T1]{fontenc}    
\usepackage{hyperref}       
\usepackage{url}            
\usepackage{booktabs}       
\usepackage{amsfonts}       
\usepackage{nicefrac}       
\usepackage{microtype}      
\usepackage{xcolor}         
\usepackage{wrapfig}

\usepackage{graphicx}	
\usepackage{amsmath}	
\usepackage{bm}         
\usepackage{booktabs}   

\usepackage{xcolor}
\usepackage{soul}

\newcommand{\wscale}{\ensuremath{\lambda}}
\newcommand{\conv}{\ensuremath{\star}}
\newcommand{\scatprop}{\ensuremath{{{U}}}}
\newcommand{\scatcoeff}{\ensuremath{{{S}}}}

\title{Simulation-based inference with scattering representations: scattering is all you need}

\author{%
  Kiyam Lin \\
  Mullard Space Science Laboratory, University College London, Holmbury St Mary, \\
  Dorking, Surrey RH5 6NT, UK \\
  Department of Physics and Astronomy, University College London, Gower Street, \\
  London, WC1E 6BT, UK \\
  \texttt{kiyam.lin@ucl.ac.uk} \\
  \AND
  Benjamin Joachimi \\
  Department of Physics and Astronomy, University College London, Gower Street, \\
  London, WC1E 6BT, UK \\
  \texttt{b.joachimi@ucl.ac.uk} \\
  \AND
  Jason D.~McEwen \\
  Mullard Space Science Laboratory, University College London, Holmbury St Mary, \\
  Dorking, Surrey RH5 6NT, UK \\
  \texttt{jason.mcewen@ucl.ac.uk} \\
}

\begin{document}
\maketitle

\begin{abstract}
	We demonstrate the successful use of scattering representations without further compression for simulation-based inference (SBI) with images (i.e.\ field-level), illustrated with a cosmological case study.  Scattering representations provide a highly effective representational space for subsequent learning tasks, although the higher dimensional compressed space introduces challenges. We overcome these through spatial averaging, coupled with more expressive density estimators. Compared to alternative methods, such an approach does not require additional simulations for either training or computing derivatives, is interpretable, and resilient to covariate shift.  As expected, we show that a scattering only approach extracts more information than traditional second order summary statistics.
\end{abstract}



\vspace{-0.2cm}
\section{Introduction}

Simulation-based inference (SBI) has recently emerged as a powerful approach for scientific inference that frees the task of parameter inference from defining a tractable and accurate likelihood \cite{cranmer2020frontier}. The premise of neural density approaches to SBI is to make use of forward simulations to learn a density estimator representing the likelihood \cite{papamakarios2019sequential}, posterior \cite{papamakarios2016fast, greenberg2019automatic} or likelihood-to-evidence ratio \cite{durkan2020proceedings, cole2022fast}.
The benefits of SBI mean that as long as we are capable of performing accurate stochastic forward simulations, it becomes possible to infer an unbiased posterior regardless of the complexity of the data being simulated.  Nevertheless, compression is a critical component of SBI for many problems and requires further study, which is the focus of the current article.  Throughout, we consider an illustrative case study from the field of cosmology involving non-Gaussian random fields.

In many scientific fields, traditional parameter inference approaches assume a Gaussian likelihood of second order statistics and simple parametric modelling of systematics to simplify the task of statistical inference \cite{asgari2021kids, abbott2022dark, joachimi2021kids}. Such an approach, however, is in many cases not sufficient to fully exploit the richness of upcoming data \cite{lin2023simulation,mvwk2024kids} and it becomes important to make use of summary statistics that contain beyond second order information.

For image data, one such approach is through the use of higher order statistics \cite[e.g.][]{pyne2022three,burger2023kids,harnois2021cosmic, munshi2020estimating, munshi2021morphology, munshi2022weak, munshi2023position, krause2024parameter}.  These statistics, however, are difficult to compute and it is often even more difficult to accurately characterise their covariance structure for use with an analytical likelihood.
Another direction involves the use of image-level (commonly called field-level) inference \cite[e.g.][]{porqueres2022lifting, boruah2022map, tsaprazi2022field, andrews2023bayesian, porqueres2023field, nguyen2024much, krause2024parameter}. Such approaches, however, must overcome the challenges of high dimensional Markov chain Monte Carlo (MCMC) sampling and approximate likelihoods.  SBI has recently been demonstrated as an effective alternative for field-level inference, eliminating these challenges \cite{jeffrey2024dark, gatti2024dark, lemos2023simbig}.
However, due to the high dimensionality of field-level data, there is a need for substantial data compression.

There are currently two main approaches to performing data compression for field-level SBI in cosmology.
One route makes use of statistical compression methods, such as MOPED \cite{heavens2000massive} or score compression \cite{alsing2018generalized}. These methods are designed such that the compression can be lossless with respect to the Fisher information.
Statistical compression methods have implementation challenges, however, as they require simulations to be run at perturbed parameters so that derivatives can be computed by finite differences.  Some simulation suites support this (allocating computational budget for this purpose) but most do not. Alternatively, automatic differentiation could be leveraged to compute derivatives \cite{campagne2023jax}, although that requires codes implemented within a framework that provides automatic differentiation, which are not always readily available.
Furthermore, statistical compression methods are not generally lossless and only asymptotically lossless around a maximum likelihood estimation point.
An alternative avenue is to make use of neural compression, of which convolutional neural networks (CNNs) have proven to be a popular architecture for field-level data \cite{ribli2019weak, jeffrey2024dark}. The drawback of neural compression is that it requires a substantial additional simulation budget for training.  Furthermore, it is not interpretable and inference can suffer from covariate shift if the training data does not match the distribution of real data.
Hybrid statistical and neural compression techniques have also been considered, such as Information Maximising Neural Networks (IMNN) \cite{charnock2018imnn}. While this realizes the advantages of both frameworks, it also realizes all of the drawbacks of both, requiring access to numerical derivatives, a significant simulation budget for generating training data, and can suffer covariate shift.

Wavelet scattering representations provide an alternative summary statistic \cite{mallat2012group, bruna2013invariant}. Whilst scattering representations are inspired by CNNs, they are designed and not learnt. This has the distinct advantage of not being a neural method.  Consequently, they are interpretable \cite{cheng2020new, allys2020new, cheng2021quantify} and do not require any training data. Scattering representations have gained traction in cosmology recently and appear to be highly effective at capturing informative non-Gaussian information \cite{cheng2020new, allys2020new, cheng2021quantify, valogiannis2024precise, cheng2024cosmological}.  In the asymptotic case of infinite depth they capture all information content of the field analysed, while also capturing the majority of information in a relatively small number of levels \cite{mallat2012group, mcewen2022scattering, cheng2021quantify}.  Furthermore, scattering transforms have recently been considered in cosmology as a form of compression for SBI \cite{gatti2024dark,blancard2023rm}.  However, they have either been combined with neural compression \cite{gatti2024dark} or demonstrated to provide unstable inferences \cite{blancard2023rm}.

In this work we demonstrate the successful use of scattering representations without further compression as summary statistics of non-Gaussian cosmological random fields for accurate SBI.  This approach does not suffer from the drawbacks of other statistical and neural compression techniques, i.e.\ it does not require additional simulations for computing derivatives, it does not require any simulations for training, is interpretable and not sensitive to covariate shift during data compression.

\vspace{-0.2cm}
\section{Wavelet Scattering Representation}

Wavelet scattering representations \cite{mallat2012group, bruna2013invariant} leverage stable wavelet representations \cite{mallat1999wavelet} to create powerful hierarchical representational spaces that satisfy isometric invariance up to a particular scale, are stable to diffeomorphisms and probe all content of the underlying field. Consequently, they provide an excellent representational space for subsequent learning \cite{bruna2011classification,mcewen2022scattering}.  In this work we consider scattering representations as the compressed representational space on which to perform neural likelihood estimation \cite{papamakarios2019sequential}.

The scattering propagator of field $f$ with wavelet $\psi_\wscale$, for parameters $\wscale=(j,\chi)$ with scale $j$ and orientation $\chi$, is given by
\begin{equation}
	\scatprop[\wscale] f
	= \vert f \conv \psi_\wscale \vert ,
\end{equation}
where $\conv$ denotes the convolution operator and the modulus function is adopted to introduce non-linearity since it is non-expansive and thus preserves the stability of wavelet representations.
Scattering representations for the path $p = (\wscale_1, \wscale_2, \ldots, \wscale_d)$, with depth $d$, can then be constructed by applying a cascade of propagators:
\begin{equation}
	\scatprop[p] f
	= \scatprop[\wscale_1, \wscale_2, \ldots, \wscale_d] f
	= \scatprop[\wscale_d] \ldots \scatprop[\wscale_2] \scatprop[\wscale_1] f
	= \vert\vert\vert f \conv \psi_{\wscale_1} \vert \conv \psi_{\wscale_2} \vert \ldots \conv \psi_{\wscale_d} \vert
	.
\end{equation}
By adopting carefully designed filters, i.e.\ wavelets, combined with a non-expansive activation function, scattering propagators inherit the stability properties from the underlying wavelets, yielding representations that are stable to diffeomorphisms.
Scattering propagators themselves, however, do not yield representations that exhibit isometric invariances.  The introduction of a form of averaging is necessary to ensure isometric invariance.
Scale-dependent averaging is performed by convolution with a scaling function $\phi$ to yield the scattering representation \cite{mallat2012group, bruna2013invariant}
\begin{equation}
	\scatcoeff[p] f
	= \vert\vert\vert f \conv \psi_{\wscale_1} \vert \conv \psi_{\wscale_2} \vert \ldots \conv \psi_{\wscale_d} \vert \conv \phi
	.
\end{equation}
For compression, scattering representations are more challenging to work with than aforementioned neural and statistical compression due to their relatively higher dimensional representational space.  We perform spatial averaging \cite[cf.][]{cheng2020new} to further compresses the representational space.

\vspace{-0.2cm}
\section{Methodology}

\textbf{Simulations:}
We adopt the high resolution Quijote $\Lambda$CDM N-body simulations \cite{villaescusa2020quijote}, extracting 20 fields per simulation by slicing each simulation cube depth-wise and making use of all 2,000 dark matter only simulations run over a latin-hypercube.
The non-linear and correlated cosmological parameters, including the cold dark matter density $\Omega_{\mathrm{m}}$, baryonic matter density $\Omega_{\mathrm{b}}$, scalar spectral index $n_{\mathrm{s}}$,  Hubble parameter $h$, and the root-mean-square matter fluctuation $\sigma_8$, are varied over the ranges shown in Appendix~\ref{appendix} (Table~\ref{tab:quijote_params}).
For this simulation suite numerical derivatives are not available, hence it is not possible to perform statistical compression (MOPED or score) or hybrid compression (IMNN).  As we are not applying any neural based compression schema, we can use all simulations for training the neural likelihood density estimator without the need to reserve any to train a compressor.

\textbf{Compression:}
To calculate the compressed scattering representation we adopt the \texttt{kymatio} package\footnote{\url{https://github.com/kymatio/kymatio}} \cite{kymatio2020} using a Morlet mother wavelet, introduced for astrophysics in \cite{mcewen2006high}, that is rotated and dilated to recover the set of wavelets considered. Since dark matter fields are expected to be isotropic on large scales (due to the cosmological principle), we set the number of spatial orientations probed to two and take the average.  We take randomly selected fields from the simulation cube sliced in depth with resolution of $512^2$ pixels and only probe scales, $J$, up to $2^{J+1} = 512$. This reduces the dimensionality of the summary statistic to a data vector of length $n = 1 + J + \frac{1}{2} J \times (J-1)=37$.
For comparison, we consider an alternative simple compression, which also does not require derivatives or training, based on binning the power spectrum of the field. This bandpower approach, however, only captures second order information.  We first calculate the power spectrum using the \texttt{pylians} package\footnote{\url{https://github.com/franciscovillaescusa/Pylians3}}, with 362 bins, and further bin into bandpowers using log spaced bins with the binning scheme described in \cite{brown2005cosmic}, resulting in a compressed dimension of 14.
We also consider a summary statistic where scattering coefficients are concatenated with bandpowers.

\textbf{SBI:}
We perform neural likelihood estimation (NLE) \cite{papamakarios2019sequential} using the \texttt{PyDELFI} package\footnote{\url{https://github.com/justinalsing/pydelfi}} \cite{alsing2019fast} with the compressed summary statistics described above.
Whilst the performance of neural posterior estimation \cite{greenberg2019automatic} may be considered in a future study, for cosmology, NLE invaluably gives direct access to a learned likelihood that can be tested such as with a goodness of fit test.
We follow the approach presented in \cite{lin2023simulation, mvwk2024kids} where an ensemble of Masked Autoregressive Flows (MAFs; \cite{papamakarios2017masked}) are considered, each with differing numbers of Masked Autoencoder for Distribution Estimation (MADE) components \citep{germain2015made}, and then model averaged, weighted by the relative validation likelihood.
We found that standard \texttt{PyDELFI} settings for the MAFs are not sufficiently expressive when learning our relatively high dimensional data vector.  Instead, we consider MAFs with four hidden layers of 50 neurons each.
Finally, we sample the posterior defined in terms of the learned likelihood using the \texttt{nautilus} package\footnote{\url{https://github.com/johannesulf/nautilus}} \cite{lange2023nautilus}.

\vspace{-0.2cm}
\section{Results}

\begin{figure}
	\begin{minipage}[c]{0.51\columnwidth}
		\centering
		\includegraphics[width=\columnwidth]{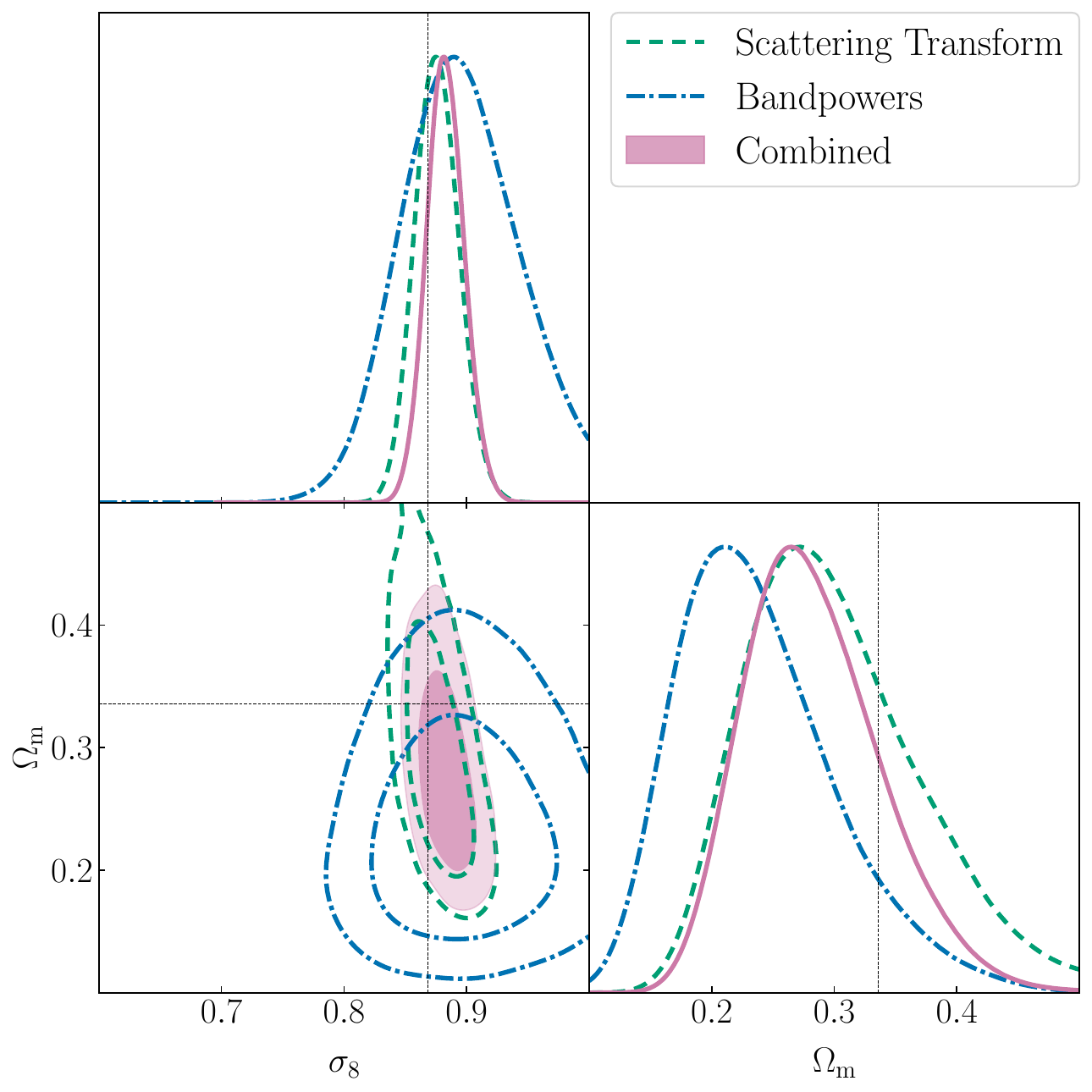}
		\caption[width=0.48\columnwidth]{Posterior constraints for different compressed representations.  Scattering representations, without further neural compression, are highly effective when using more expressive density estimators.}
		\label{fig:main_result}
	\end{minipage}
	\hspace{0.2cm}
	\begin{minipage}[c]{0.46\columnwidth}
		\centering
		\includegraphics[width=\columnwidth]{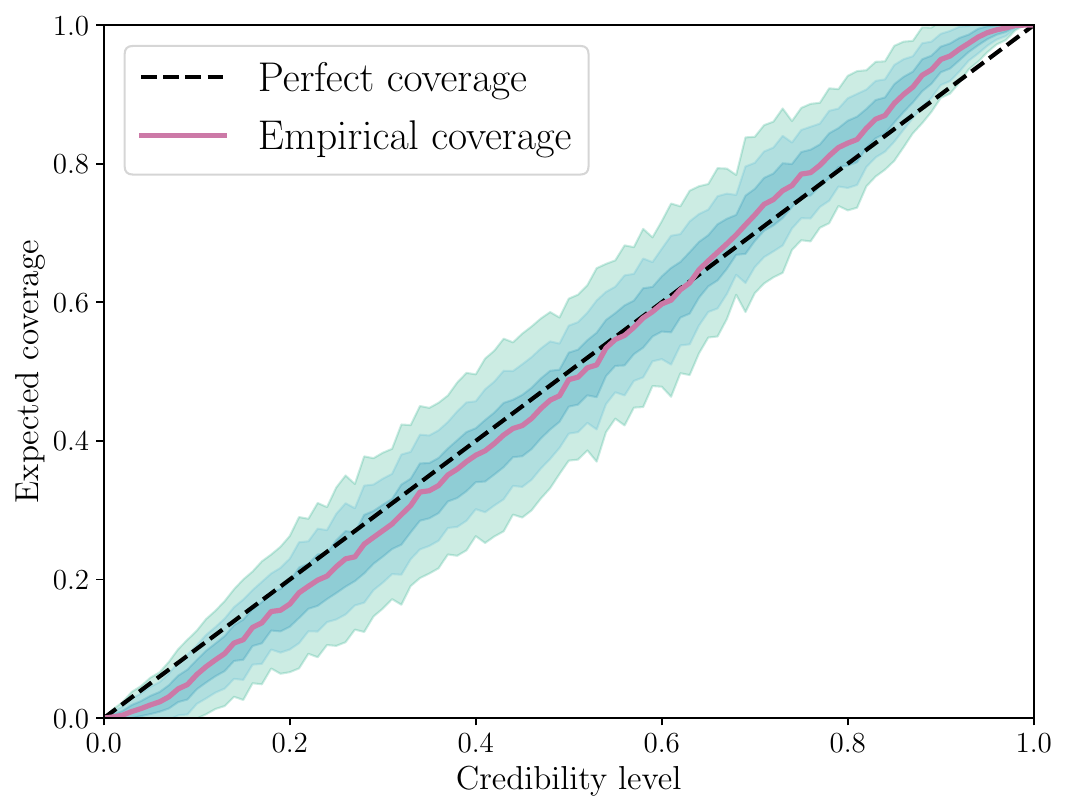}
		\caption{Coverage tests using TARP \cite{lemos2023sampling} for the combined scattering and bandpower summary statistic (similar results are found when considering each summary statistic separately). The blue regions decreasing in saturation indicate the 1, 2 and 3 $\sigma$ errors obtained from bootstrapping over 100 realisations. The dashed black line denotes perfect coverage. Empirical coverage is generally good. }
		\label{fig:combined_coverage}
	\end{minipage}
	\vspace{-3mm}
\end{figure}

Performing SBI on a mock data vector (parameters in Table~\ref{fig:main_tab}), we find that SBI with scattering performs well without further compression as indicated by the results of Figure~\ref{fig:main_result} and Table~\ref{fig:main_tab} (on a single CPU core with 8GB RAM). We only show $\Omega_{\mathrm{m}}$ and $\sigma_8$ results as the other three cosmological parameters are not well constrained by dark matter only N-body simulations, as expected.  Inspection of the ensemble of trained MAFs demonstrates the density estimators are converged, in contrast with cases considered in \cite{blancard2023rm}.
For comparison, bandpowers are less effective as they only capture second order statistical information.  Since the scattering and bandpower representations capture different statistical information, constraints can be improved by combining both summary statistics.
To quantify the improvement, we find that scattering alone can yield $\sim$60\% tighter constraints in $\sigma_8$ than bandpowers alone, boosting the improvement to $\sim$70\% in the combined summary statistics case.

\begin{table}
	\centering
	\begin{tabular} {l c c c c}
		\toprule
		Parameter             & Truth & Scattering                & Bandpowers                & Combined                  \\
		\midrule
		{$\sigma_8$}          & 0.869 & $0.876\pm 0.018$          & $0.894\pm 0.046$          & $0.883\pm 0.015$          \\
		{$\Omega_\mathrm{m}$} & 0.336 & $0.298^{+0.052}_{-0.083}$ & $0.238^{+0.045}_{-0.077}$ & $0.283^{+0.044}_{-0.064}$ \\
		\bottomrule
	\end{tabular}
	\caption{Posterior constraints on $\sigma_8$ and $\Omega_{\mathrm{m}}$ with one standard deviation errors for different compressed representations.}
	\label{fig:main_tab}
\end{table}

To ensure that the posterior distribution we infer is representative and unbiased, we conduct coverage tests based on the Tests of Accuracy with Random Points (TARP) algorithm \cite{lemos2023sampling}. TARP measures the expected coverage probability for a given credibility level empirically from samples. Results are shown in Figure~\ref{fig:combined_coverage}, where the empirical coverage is shown to approximate the ideal perfect coverage scenario resonably well. Here we consider the combined summary statistic, although we perform the same test on all considered summary statistics and find good coverage, although with slight under confidence, which is conservative.

\vspace{-0.2cm}
\section{Conclusions}

We demonstrate that field-level SBI can be performed effectively using wavelet scattering representations alone to provide a compressed representation without any further neural compression.  The challenge of a higher dimensional compressed representation is overcome by incorporating spatial averaging and a judicious choice of parameters, coupled with more expressive density estimators.  This approach has considerable advantages compared to the alternative statistical or neural compression approaches typically considered in that it does not require additional simulations to train a neural compressor, it does not require simulation suites with perturbed parameters needed to compute numerical derivatives for statistical compression approaches, it is more interpretable than the use of black-box neural networks, and it is not sensitive to covariate shift.
Furthermore, we demonstrate that SBI with scattering representations extracts beyond two-point statistical information, as expected, since recovered parameter constraints are distinct and tighter than those recovered by bandpowers (thus, scattering representations and bandpowers can be combined to improve constraining power further still).
In parallel with this work, successful use of scattering-only compression for SBI has simultaneously been demonstrated for 21cm epoch of reionization signals \cite{zhao2024simulation}, further evidencing the effectiveness of scattering for SBI.
In a sense, for field-level SBI, scattering is all you need for compression.  That said, we do not dogmatically advocate that scattering alone should always be used.  While the scattering approach presented has numerous advantages, our findings motivate further studies directly comparing the effectiveness of different compression approaches for field-level SBI.

\begin{ack}

	The authors would like to thank Matt Price and Alessio Spurio Mancini for invaluable discussions. KL thanks the UCL Cosmoparticle Initiative and STFC (grant number ST/W001136/1). BJ acknowledges support by the ERC-selected UKRI Frontier Research Grant EP/Y03015X/1 and by STFC Consolidated Grant ST/V000780/1. JDM is supported by EPSRC (grant number EP/W007673/1) and STFC (grant number ST/W001136/1). This work used facilities provided by the UCL Cosmoparticle Initiative. We thank Dr Edd Edmonson for technical support at the UCL HPC facilities.

\end{ack}



\bibliographystyle{abbrvnat}
\bibliography{scattering_bib} 

\newpage



\appendix

\vspace{-0.2cm}
\section{Cosmological parameter ranges}
\label{appendix}

Table~\ref{tab:quijote_params} shows the priors ranges used both for the Quijote N-body simulation latin-hypercube and for inference. For inference, uniform priors were adopted for all parameters that matched the prior width of the simulations such that our learnt likelihood had full training data support.

\begin{table}[h]
	\centering
	\begin{tabular}{l c}
		\toprule
		Parameter             & Prior range    \\
		\midrule
		$\Omega_{\mathrm{m}}$ & \ [0.1, 0.5]   \\
		$\Omega_{\mathrm{b}}$ & \ [0.03, 0.07] \\
		$h$                   & \ [0.5, 0.9]   \\
		$n_{\mathrm{s}}$      & \ [0.8, 1.2]   \\
		$\sigma_8$            & \ [0.6, 1.0]   \\
		\bottomrule
	\end{tabular}
	\caption{Cosmological parameter ranges.}
	\label{tab:quijote_params}
\end{table}

\end{document}